\documentclass[conference]{IEEEtran}
\IEEEoverridecommandlockouts
\usepackage{cite}
\usepackage{amsmath,amssymb,amsfonts}
\usepackage{algorithmic}
\usepackage{graphicx}
\usepackage{textcomp}
\usepackage{xcolor}
\usepackage{subfig}
\usepackage[font=normalsize,labelfont=bf]{caption}
\usepackage[
  separate-uncertainty = true,
  multi-part-units = repeat
]{siunitx} 

\def\BibTeX{{\rm B\kern-.05em{\sc i\kern-.025em b}\kern-.08em
    T\kern-.1667em\lower.7ex\hbox{E}\kern-.125emX}}
\begin{document}

\title{Prediction of Overall Survival of Brain Tumor Patients\\
}

\author{\IEEEauthorblockN{Rupal R. Agravat}
\IEEEauthorblockA{\textit{School of Engineering and Applied Science,} \\
\textit{Ahmedabad University and } \\
\textit{Institute of Technology, Nirma University}\\
\textit{Ahmedabad, India} \\
\textit{rupal.kapdi@nirmauni.ac.in}}
\and
\IEEEauthorblockN{Mehul S. Raval}
\IEEEauthorblockA{\textit{School of Technology,} \\
\textit{ Pandit Deendayal Petroleum University,}\\
\textit{Gandhinagar, India} \\
\textit{mehul.raval@sot.pdpu.ac.in}}
\and
}
\maketitle

\begin{abstract}
Automated brain tumor segmentation plays an important role in the diagnosis and prognosis of the patient. In addition, features from the tumorous brain help in predicting patients' overall survival. The main focus of this paper is to segment tumor from BRATS 2018 benchmark dataset and use age, shape and  volumetric features to predict overall survival of patients. The random forest classifier achieves overall survival accuracy of 59\% on the test dataset and 67\% on the dataset with resection status as gross total resection. The proposed approach uses fewer features but achieves better accuracy than state-of-the-art methods.
\end{abstract}


\section{Introduction}
Medical fraternity considers brain tumor amongst the most fatal type of cancer\cite{b2}. Brain tumors are divided into two categories based on origin and malignancy. Former is further classified as primary and secondary. The primary tumor develops in the brain whereas secondary spreads from another body part to the brain. According to the World Health Organization (WHO), malignancy based tumors can be classified in grades I to IV according to increasing aggressiveness~\cite{b3}. High-Grade Glioma (HGG) (grade III and grade IV tumor), needs immediate treatment~\cite{b3}. It may lead to patient's death in less than two years, whereas Low-Grade Glioma (LGG) is the benign tumor which grows slowly and the patient has several years of life expectancy.

Magnetic Resonance Imaging (MRI) is a preferred technique for capturing tumors in the brain as it provides good soft tissue contrast \cite{b4}. MRI sequences are also acquired by injecting Gadolinium to enhance and improve the quality of the MRI images\cite{b7}. Usually, human expert uses MRI images for the tumor diagnosis. The task is quite challenging due to the large data volume\cite{b8}. This motivates the need for automated or semi-automated brain tumor segmentation. Automated brain tumor segmentation is divided into three categories: basic, generative, and discriminative \cite{b14},\cite{b34}. With the evolution of deep learning, state-of-the-art methods use Convolution Neural Network(CNN) for semantic segmentation of the tumor~\cite{b19}.

Many methods further segment the tumor into its substructures like; necrosis, enhancing tumor and edema. Size of the tumor and size of substructures play a major role in predicting the overall survival (OS). In \cite{b16}, 3D U-net based model for tumor segmentation and radiomics based features are used for overall survival prediction. The tumor is characterized by image-based features computed from the segmentation masks. These features are then used to train a Random Forest Regressor (RFR) with 1000 trees and ensemble of small multilayer perceptrons (MLP). The reported accuracy is 52.6\% on the test dataset  for overall survival and the Spearman correlation coefficient of 0.496.  

In another attempt at survival prediction~\cite{b17}, the authors use pre-trained AlexNet to segment the brain tumor. The features from segmentions are used to train the linear discriminant for survival prediction. The texture features resulted in the accuracy of 46\%, and histogram features achieved an accuracy of 68.5\% for the test dataset. The authors developed a fully automated model for segmentation of LGG and HGG in multimodal MRIs\cite{b18}. The prediction of patient overall survival is based on support vector machine (SVM) learning algorithms. They reported 100\% accuracy for overall survival prediction on a set of 16 test samples. In \cite{b33}, authors use Dense-Res-Inception Net(DRINet) for biomedical image segmentation. The paper reported 83.47\%, 73.41\%, 64.98\% Dice Similarity Coefficient(DSC) for whole tumor, tumor core and enhancing tumor respectively.  

In \cite{b20}, a fully convolution neural network(FCNN) architecture is used for tumor segmentation and the extracted features are fed to SVM classifier for OS prediction. A preprocessing step on MR scans is done using Z-score normalization to overcome multi-center data and magnetic field inhomogeneities. Also, post-processing is implemented using connected components to remove components below the threshold. The features are extracted from segmented regions and fed to SVM with a linear kernel. The reported accuracy for OS prediction is 60\%. In ~\cite{b23} authors created an ensemble using 19 variations of DeepMedic and 7 variations of 3D U-net. Various features namely age, spatial, volumetric, morphological and tractographic are extracted, and their combination is used to train the SVM classifier. The authors reported an accuracy of 70\% for features from ground truth and 63\% for features from network segmentation. Both the accuracies are reported on the data of 59 patients with resection status as gross total resection(GTR). 

In~\cite{b24}, authors implemented DeepMedic CNN architecture for tumor segmentation and implemented the cox model for OS prediction. They achieved 80\%, 68\% and 67\% DSC for whole tumor, tumor core and enhancing tumor respectively. The OS prediction accuracy for training dataset is 44.5\% and for test dataset is 38.2\%. 

Authors in~\cite{b25}, implemented PixelNet architecture for tumor segmentation and achieved 88\% whole tumor DSC. The artificial neural network (ANN) is trained on mean, skewness, and location of tumor for OS prediction. They reported an accuracy of 54.5\%. In~\cite{b26}, the authors implemented densely connected convolution neural network for segmentation, and MLP based regressor for OS prediction. They reported 50\% accuracy for training data. Authors in~\cite{b27} implemented the ensemble of three convolution networks with hybrid loss function and extracted radiomics features to train random forest classifier. They reported accuracy of 52.6\% on the validation set. In~\cite{b28}, authors modified U-net architecture with bottleneck layers and dense layers and applied elastic net for OS prediction. They reported an accuracy of 67\% for the training data.

Authors in~\cite{b29} implemented extended U-net architecture for tumor segmentation and XGBoost regression for OS prediction. They achieved 65\% accuracy on training data. In ~\cite{b30}, residual U-net is implemented for tumor segmentation. The ensemble of regression network and random forest classifier is used for survival prediction. The paper reported accuracy of 47.5\%.

The above methods either uses segmentation model with large number of network parameters or use more features for training the classifier. The literature suggests that U-net architecture provides good semantic segmentation. Therefore, the paper uses U-net architecture as proposed in ~\cite{b21} with modifications. The proposed work reduces network depth to minimize network parameters. The inductive transfer learning~\cite{b32} is used for substructures segmentation. The whole tumor segmentation weights are transferred to the networks which train substructures. The weight transfer has substantially reduced the problem of training failure and it allows network to learn from small annotated data. The volumetric and shape features are extracted from the segmented results.  Along with these features, age is used to train random forest classifier for OS prediction.

This paper is organized as follows. Section II presents preliminaries about the brain tumor segmentation (BraTS)~\cite{b19} dataset used in the proposed work. Section III covers the CNN used for brain tumor segmentation and random forest classifier for overall survival prediction. Section IV discusses the experimental results. Finally, Section V concludes the paper, with suggestions to further improve the OS prediction.

\section{Multimodal Brain Tumor Segmentation Challenge}
\begin{figure}[!h]
\centering
   \includegraphics[width=0.5\textwidth]{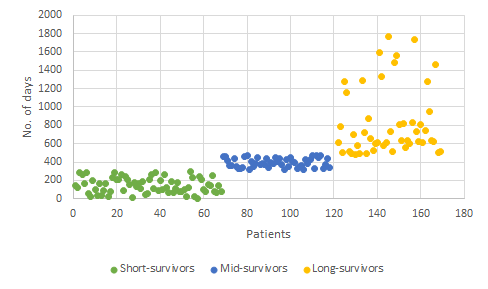} 
 \caption{Patients' OS days.}
\label{Survival Graph}
\end{figure}

The multimodal brain tumor segmentation challenge invites researchers to develop robust brain tumor segmentation techniques from MRI scans~\cite{b11,b19}. The data set handles all ethical issues with care. The BraTS 2018 challenge has two tasks: segmentation of the gliomas, and prediction of patient's OS. The dataset~\cite{b11,b12,b13,b19} comprised of clinically-acquired 3T multimodal MRI scans and all the ground truth labels were manually revised by expert board certified neuro-radiologists. Annotations are the Gd-enhancing tumor (ET — label 4), the peritumoral edema (ED — label 2), and the necrotic and non-enhancing tumor (NCR/NET — label 1) ~\cite{b19}. The dataset is co-registered to the same anatomical template, interpolated to the same resolution (1$mm^{3}$) and skull-stripped~\cite{b21}. The dataset has 210 HGG samples and 75 LGG samples, with each sample having four MRI modalities ($ T_{1}$, $T_{2}$, $T_{1}C$, and FLAIR) along with the ground truth. Each sample has 155 slices with 240x240 pixels per slice. Features related to patients' OS are also provided like the number of survival days, resection status (GTR / sub total resection(STR)) and the age. The suggested classes based on the prediction of OS were long-survivors ($>$15 months), short-survivors ($<$10 months), and mid-survivors (between 10 to 15 months). Overall survival of patients for number of days is shown in Fig. \ref{Survival Graph}. Age and OS days distribution among three survival classes is shown in Table \ref{age days range}. The number of short survivors are higher compared to other classes. The mean age of such patients is also high and median age is 66.55. In addtion, their OS days are less in comparison to other classes. Whereas long-survivors are less in number but they have higher OS span compared to other two classes. One can also observe high variability in data of the long-survivors. 

\begin{table}[htbp]
\centering
\caption{Distribution of dataset features in survival classes.}
\label{age days range}

\begin{tabular}{|c|p{1.5 cm}|c|c|}
\hline
\textbf{Survival class} & \textbf{\# Patients} & \textbf{Age ($\mu$ $\pm$ $\sigma$)} & \textbf{OS days ($\mu$ $\pm$ $\sigma$)} \\
\hline
Short-survivors & 65 & 65.44 $\pm$ 10.68 & 147.44 $\pm$ 83.08 \\
\hline 
 Mid-survivors & 50 & 58.70 $\pm$ 11.26 & 394 $\pm$ 49.32  \\
\hline
 Long-survivors & 48 & 55.11 $\pm$ 12.19 & 826.23 $\pm$ 370.91 \\
\hline

\end{tabular}

\end{table}

\section{Implementation Details}
U-net architecture promises good semantic segmentation for biological images as shown in~\cite{b21}. Authors in~\cite{b22} also adopted the U-net architecture for brain tumor segmentation for 2D images. The proposed work considers the architecture of ~\cite{b21-b22} with minor modification. The proposed work uses three down-sampling and two up-sampling modules in the network instead of five down-sampling and four up-sampling modules of~\cite{b21} and~\cite{b22}. It is found that reduction in network depth reduces number of parameters, speeding up the processing without compromising the accuracy. Each up/down sampling module has two convolution layers. Relu activation function is applied after convolution operation and Dice loss function is used to calculate network loss after each epoch. The network is trained on whole tumor as well as on each substructure i.e., edema, enhancing tumor and necrosis.

\begin{figure}[!h]
\centering
   \includegraphics[width=0.5\textwidth]{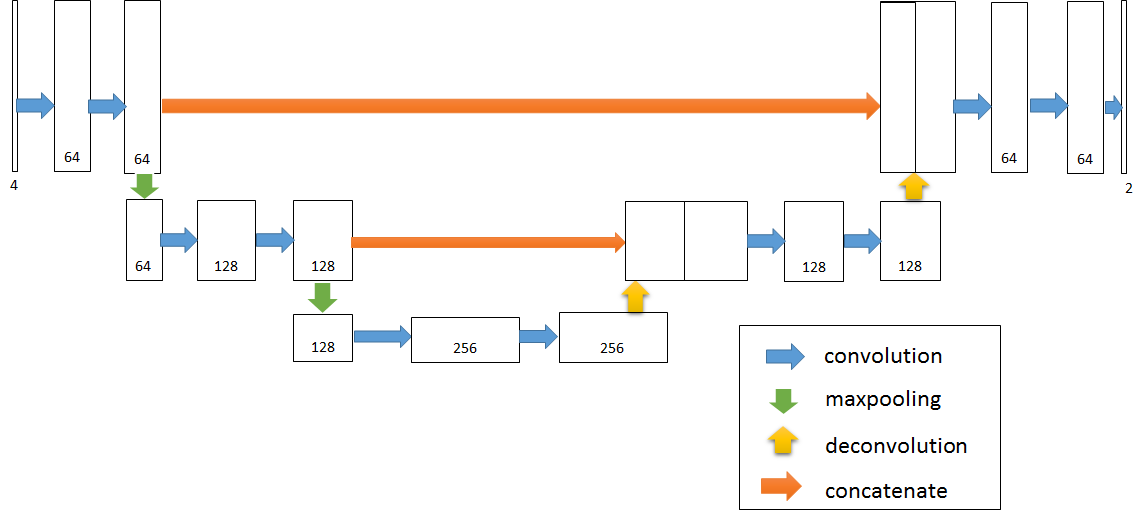} 
 \caption{U-net architecture with reduced depth.}
\label{U-net}
\end{figure}

Fig. \ref{U-net} shows the architecture used for tumor segmentation in this paper. The labeled data is highly imbalance as negative class dominates positive class. For e.g., if the whole tumor spread cover 30\% slices, then necrosis and enhancing tumor spread is found only in 10\% of the brain slices. This reduction in the positive class, makes the network training difficult. This may result in network being stuck to local minima which requires re-initiation of the network training. Such training also results in large amount of false positives. The concept of inductive transfer learning as suggested in ~\cite{b32} helps to resolve the issue. Source domain (whole tumor) network parameters are transferred to target domain (substructure) network and these parameters are used to initialize the target network training. The parameter transfer serves three purposes: 1) It deals with scarcity of labelled data; 2) provides localization for substructure area and 3) reduces amount of false positives. Weight transfer has improved the network training performance. Data preprocessing includes Z-score normalization and data augmentation by applying rotation, flipping, elastic transformation, shear, shift and zoom to the MRI slices.

Following features are extracted from the whole tumor and substructures for training random forest classifier:

\begin{itemize}
  \item \textbf{Volumetric features} include the volume of the tumor with respect to the brain, the volume of necrosis, edema, enhancing tumor with respect to the whole tumor, and extent of the tumor.
  
  \item \textbf{Shape features} include elongation, flatness, minor axis length, major axis length, maximum 2D diameter, maximum 3D diameter, mesh volume, sphericity.
\end{itemize}

Five volumetric features, fourteen shape features, and age is used to train random forest classifier with 5-fold cross-validation. 

\section{Experimental Results}
The work uses NVIDIA Quadro K5200 and Quadro P5000 GPUs for training and testing of CNN and random forest training. Python 3.6 along with all the necessary packages is used for development. The U-net is trained over 80\% (228) images and tested on 20\%(57) images. The training data is further divided into training(204 images) and validation(24 images) sets. 

\begin{figure}[!h]
\centering

\subfloat[\label{fig1}]{%
   \includegraphics[width=0.12\textwidth]{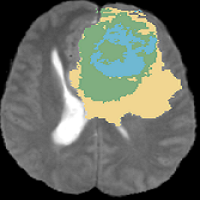}
 }
  \hfill
\subfloat[\label{fig2}]{%
   \includegraphics[width=0.12\textwidth]{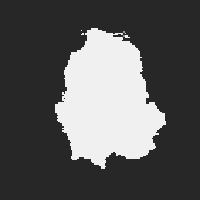}
 }
 \hfill
\subfloat[\label{fig3}]{%
   \includegraphics[width=0.12\textwidth]{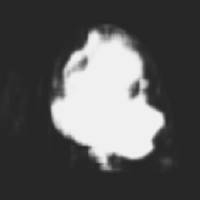}
   }
   
  \vfill
\subfloat[\label{fig4}]{%
   \includegraphics[width=0.12\textwidth]{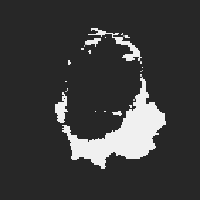}
 }
  \hfill
\subfloat[\label{fig5}]{%
   \includegraphics[width=0.12\textwidth]{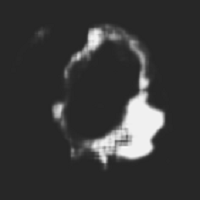}
 }
 \hfill
\subfloat[\label{fig6}]{%
   \includegraphics[width=0.12\textwidth]{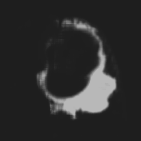}
 }
 
 \vfill
\subfloat[\label{fig7}]{%
   \includegraphics[width=0.12\textwidth]{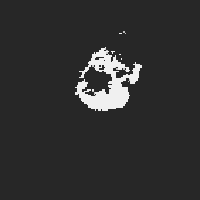}
 }
  \hfill
\subfloat[\label{fig8}]{%
   \includegraphics[width=0.12\textwidth]{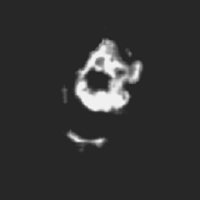}
 }
 \hfill
\subfloat[\label{fig9}]{%
   \includegraphics[width=0.12\textwidth]{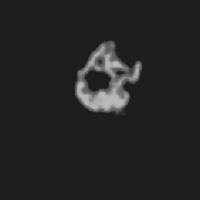}
 }
 
\vfill
\subfloat[\label{fig10}]{%
   \includegraphics[width=0.12\textwidth]{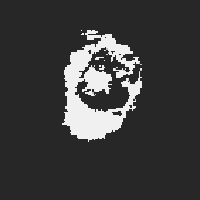}
 }
  \hfill
\subfloat[\label{fig11}]{%
   \includegraphics[width=0.12\textwidth]{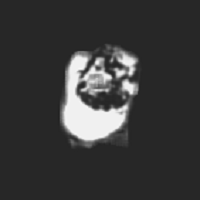}
 }
 \hfill
\subfloat[\label{fig12}]{%
   \includegraphics[width=0.12\textwidth]{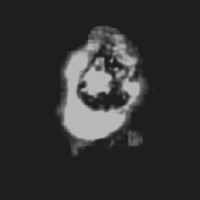}
 }

 \caption{Segmentation results: a) T2 image, (where, yellow represents Edema, blue represents Enhancing Tumor, and green represents Necrosis/Non-enhancing tumor) b) Whole tumor ground truth, c) Whole tumor segmentation, d) Edema ground truth, e) Edema segmentation without weight initialization, f) Edema segmentation with weight initialization, g) Enhancing tumor ground truth, h) Enhancing tumor segmentation without weight initialization, i) Enhancing tumor segmentation with weight initialization, j) Necrotic ground truth, k) Necrotic segmentation without weight initialization, l) Necrotic segmentation with weight initialization.}
\label{CNN-Res}
\end{figure}

The dataset is highly imbalanced as tumor occupies small portion of the brain. The substructures of the tumor occupy an even smaller volume compared to the whole tumor. Initially, the network is trained to segment whole tumor with initialization of parameters randomly chosen from normal distribution. The obtained weights inturn are transferred to the substructure networks for parameter initialization. During each run network is trained for 50 epochs. Fig. \ref{CNN-Res} shows segmentation results for the whole tumor with three substructures(with and without inductive transfer learning) for a sample 2D slice.

Table \ref{DSC} shows the dice similarity coefficient, sensitivity and predictive positive rate(PPV) for the test dataset of 57 patients (HGG - 42, LGG - 15). One can observe that inductive transfer learning improves segmentation result.

\begin{table}[htbp]
\centering
\caption{DSC, sensitivity and positive predictive value (PPV) for test dataset, A: without weight transfer, B: with weight transfer.}
\label{DSC}
\begin{tabular}{|p{1.2 cm}|c|c|c|c|c|c|}
\hline
 & \multicolumn{2}{|c|}{\textbf{DSC}} & \multicolumn{2}{|c|}{\textbf{Sensitivity}} & \multicolumn{2}{|c|}{\textbf{PPV}}  \\
\hline
\textbf{Whole Tumor} & \multicolumn{2}{|c|}{0.78} & \multicolumn{2}{|c|}{0.76} & \multicolumn{2}{|c|}{0.91} \\
\hline
 & A & B & A & B & A & B \\
\hline 
\textbf{Necrosis} & 0.58 & \textbf{0.65} & 0.56 & \textbf{0.67} & 0.69 & \textbf{0.70} \\
\hline 
\textbf{Enhancing} & 0.58 & \textbf{0.60} & \textbf{0.56} & 0.54 & 0.69 & \textbf{0.74} \\
\hline
\textbf{Edema} & 0.63 & \textbf{0.71} & 0.56 & \textbf{0.66} & 0.79 & \textbf{0.83} \\
\hline

\end{tabular}
\end{table}

Once the model is ready, segmentation is applied to data of 163 patients whose survival expectancy is provided. Volumetric and shape based features are extracted from the segmentation results. The shape-based features are extracted using pyradiomics~\cite{b31}. 

Overall survival prediction using random forest classifier with five-fold cross-validation is shown in Table \ref{RF results}. The results are shown for two types of datasets, 1) test dataset (out of 163 patients' data, 130 patients' images are used for training and 33 patients' images are used for testing) 2) set with GTR status (59 patients). 

Table \ref{Comparison} compares the accuracy of the proposed method with other state-of-the-art techniques. Methods proposed in~\cite{b16,b17} uses the dataset of BraTS 2017 whereas other methods~\cite{b23, b24, b25, b26, b29, b30} use BraTS 2018 dataset. In~\cite{b23}, classifier training set is made up of the images with GTR status. Whereas, other methods uses the training set with resection status as either GTR/ STR/ NA. Better OS accuracy is achieved in the proposed method due to; 1. Use of U-net architecture with fewer parameters; 2. Inductive transfer learning for substructure segmentation.

\begin{table}[htbp]
\centering
\caption{Overall survival prediction accuracy.}
\label{RF results}

\begin{tabular}{|c|c|c|c|}
\hline
\textbf{Feature} & \textbf{Test dataset} & \textbf{GTR dataset} \\
\hline
\textbf{Age + Volumetric + Shape} & 59\% & \textbf{67\%} \\
\hline
\textbf{Age + Volumetric} & 46\% & 64\% \\
\hline 
\textbf{Shape + Volumetric} & 50\% & 65\% \\
\hline
\textbf{Age} & 31\% & 52\% \\
\hline
\end{tabular}

\end{table}

\begin{table}[htbp]
\centering
\caption{Comparison with state-of-the-art methods.}
\label{Comparison}

\begin{tabular}{|c| p{4 cm}|c|}
\hline
\textbf{Ref.} & \textbf{Classifier(s)} & \textbf{Accuracy\%} \\
\hline
\cite{b16} & Ensemble of random forest and multi layer perceptron & 52.6\\
\hline
\cite{b17} & Linear Discriminant & 46 \\
\hline
\cite{b23} & Linear SVM GTR set & 63 \\
\hline
\cite{b24} & Neural network and random forest & 38 \\
\hline
\cite{b25} & Artificial neural network & 54.5\\
\hline
\cite{b26} & Multi layer perceptron & 50.8\\
\hline
\cite{b29} & XGBoost & 65\\
\hline
\cite{b30} & Ensemble of random forest and regression network & 47.5 \\
\hline
\textbf{Proposed} & Random forest &  \textbf{59/67(GTR)}\\
\hline
\end{tabular}
\end{table}

\subsection{Feature Analysis}
It can be observed from Table \ref{RF results} that combination of age, volumetric and shape features are best suited for the model. Additionally  other features like first order, gray level co-occurrence matrix (GLCM), gray level difference matrix (GLDM), gray level run length matrix (GLRLM) can be extracted from various modalities to improve the life expectancy results. However, higher order features are not useful in the present work as they require near perfect segmentation. Though it must be noted that improvements in the segmentation results can increase the accuracy of the OS prediction. 

\section{Conclusion}
The paper uses modified U-net architecture with reduced depth of three layers. In addition, the model is trained for the whole tumor with random parameter initialization. Substructure networks are initialized using weights of the whole tumor network. After completion of the substructure network training, segmentation results are generated for test dataset. Random forest classifier is trained on the extracted features for OS prediction. The proposed work achieves better accuracy compared to state-of-the-art methods. The accuracy can be enhanced by improving segmentation. It can be further improved by refining the network or implementing post-processing on segmentation. The future work will focus on improving the segmentation and using features from MRI modalities to improve overall survival prediction.

\section*{Acknowledgment}
The authors would like to thank NVIDIA Corporation for donating the Quadro K5200 and Quadro P5000 GPU used for this research, Dr. Krutarth Agravat (Medical Officer, Essar Ltd) and Mr. Abhishek Shah (L.G. Medical College, Ahmedabad) for clearing our doubts related to medical concepts. The authors acknowledge continuous support from Professor Sanjay Chaudhary and Mr Himanshu Budhia for this work.

\end{document}